\documentclass{article}
\usepackage{ijcai23}

\usepackage{times}
\usepackage{soul}
\usepackage{url}
\usepackage[urlcolor=blue]{hyperref}
\usepackage[utf8]{inputenc}
\usepackage[small]{caption}
\usepackage{graphicx}
\usepackage{amsmath}
\usepackage{amsthm}
\usepackage{booktabs}
\usepackage[switch]{lineno}

\usepackage{graphicx}
\usepackage[utf8]{inputenc} 
\usepackage{amsfonts}       
\usepackage{nicefrac}       
\usepackage{microtype}      
\usepackage{xcolor}         

\usepackage[vlined, ruled, linesnumbered]{algorithm2e}
\usepackage{amssymb}
\usepackage{dutchcal}

\usepackage{subfigure}
\usepackage{multirow}
\usepackage{booktabs}
\usepackage{mathrsfs}
\usepackage{enumerate}

\usepackage{booktabs}
\usepackage{romannum}
\usepackage{multirow}
\usepackage{stfloats}

\newcommand{\comment}[2][.52\linewidth]{%
  \leavevmode\hfill\makebox[#1][l]{$\triangleright$ #2}}
\newcommand{\tinycomment}[2][.37\linewidth]{%
    \leavevmode\hfill\makebox[#1][l]{$\triangleright$ #2}}
\newcommand{\tinytinycomment}[2][.3\linewidth]{%
    \leavevmode\hfill\makebox[#1][l]{$\triangleright$ #2}}

\usepackage{hyperref,tablefootnote,footnotehyper}
\usepackage[T1]{fontenc}
\usepackage{caption}

\usepackage{setspace}
\newtheorem{definition}{Definition}

\title{Data Origin Inference in Machine Learning}

\author{
Mingxue Xu$^1$
\and
Xiang-Yang Li$^2$
\affiliations
$^1$Department of Computing, Imperial College London\\
$^2$School of Computer Science and Technology, University of Science and Technology of China
\emails
mx1221@ic.ac.uk,
xiangyangli@ustc.edu.cn
}

\begin{document}

\maketitle

\begin{abstract}
It is a growing direction to utilize unintended memorization in ML models to benefit real-world applications, with recent efforts like user auditing~\cite{Song2019AuditingDP,Miao2021TheAA}, dataset ownership inference~\cite{Maini2021DatasetIO} and forgotten data measurement~\cite{jagielski2022measuring}.
Standing on the point of ML model development, we introduce a process named {\it data origin inference}, to assist ML developers in locating missed or faulty data origin in training set without maintaining strenuous metadata.
We formally define the data origin and the data origin inference task in the development of ML model (mainly neural networks). Then we propose a novel inference strategy, combining embedded-space multiple instance classification and shadow training~\cite{MIA}.
Diverse use cases cover language, visual and structured data, with various kinds of data origin (e.g. business, county, movie, mobile user, text author). 
A comprehensive performance analysis of our proposed strategy contains referenced target model layers, available testing data for each origin, and in shadow training, the implementations of feature extraction as well as shadow models.  
Our best inference accuracy achieves 98.96\% in the language use case when the target model is a transformer-based deep neural network.
Furthermore, we give a statistical analysis of different kinds of data origin to investigate what kind of origin is probably to be inferred correctly.
The code is available at \url{https://github.com/Mingxue-Xu/ori}.
\end{abstract}

\section{Introduction}\label{sec:intro}

Data origin significantly impacts the performance of the deployed Machine Learning (ML) models~\cite{10.14778/3554821.3554857}.
There are normally two origin-related reasons for the deployed ML model failures: \romannum{1}) the data domain of data origin in the training set misaligns with that of the deployment environment; \romannum{2}) the training data obtained from the data origin is dirty or poisoned.
For \romannum{1}), a straightforward solution is to train the current snapshot of the ML models incrementally with new data, which is collected from the actual deployment environment.
For \romannum{2}), it is feasible to ``delete" dirty or poisoned training data from the already trained ML models, based on a recent research branch named machine unlearning~\cite{unlearning}.   
However, both solutions need to identify whether the concerned data origin is involved in the training set. 
In practice, this information is difficult to document accordingly by {\it metadata} before model training and deployment. 
Even worse, sometimes the ML model is not initially trained by the ML developer (i.e. pre-trained model), and the original training set and metadata might not be available to the ML developer.

To this end, this work proposes a data origin inference process - {\bf given the samples from a known data origin, determine whether this origin gets involved in the model's training set}. The data origin here means {\it where the data is generated or what subject the data describe}, which is also the implicit high-resolution information of the data~\cite{survey17}. The formal mathematical definition of data origin is given in Section~\ref{subsec:origin}. We also take a reasonable assumption that the exact training data samples are {\it not} required, which is fairly relaxed compared with a similar problem named Membership Inference (MI) ~\cite{MIA}.

Our work distinguishes from existing similar work, user auditing~\cite{Song2019AuditingDP,Miao2021TheAA}, from   
the relaxed definition of the targeted object - ``origin''. Their targeted object is limited to a specific application (i.e. ``users'' in text generation or automatic speech recognition).
This relaxed definition makes our work particularly suitable for identifying training data gap, when it is unclear which kind of data origin significantly impacts the model performance. 
For instance, when a business review classification model fails because of domain misalignment, for the ML developer, it is unclear if there should collect the data of more reviewers (the data generator) or more businesses (the data subject). Thus they can try these two sorts of origins and then improve the ML model accordingly.

There are two technical bedrocks to our solution. The first is shadow training, which is widely adopted in MI. 
The second is the embedded-space multiple instance classification - we use this technique to overcome not having the exact training data samples when inference, which is different from the original assumption of MI.

To summarize, our contribution is as follows:
\begin{enumerate}
    \item As far as we know, this is the first work to explore unintended feature memorization on the data origin scale. We propose an efficient inference strategy that combines shadow training and multiple instance learning.
    \item Our evaluations cover five types of data origins with image, text, and tabular data. Experimental results show that our proposed solution significantly outperforms existing sample-level MI and achieves 83.19-98.96\% accuracy.
    \item We give a thorough performance analysis regarding the referenced model layer for inference, the amount of available data for each origin when testing, feature extraction in multiple instance learning, and shadow model trained from scratch or trained incrementally from the deployed model. We also use the Pearson correlation coefficient to address what kind of data origin has better inference performance.
\end{enumerate}
\section{Related Work}

\begin{algorithm*}[t]
    \SetKwComment{Comment}{$\triangleright$\ }{}
    \SetKwInOut{Input}{Input}
    \SetKwInOut{Output}{Output}
    \SetKwInOut{Initialize}{Initialize}
    \SetKwFunction{InterSplit}{InterSplit}
    \SetKwFunction{IntraSplit}{IntraSplit}
    \SetKwFunction{TrainShadow}{TrainShadow}
    \SetKwFunction{TrainMeta}{TrainMeta}
    \SetKwFunction{SplitB}{SplitB}
    \SetKwFunction{Featurize}{Feat}
    \SetKwFunction{Part}{Partition}
    \SetKwFunction{GenData}{GenData}
    
    \Input{target model $f$ and its hyperparameters $f_{\theta}$, $i$th layer access $h_i(\cdot)$, proxy dataset $D^{proxy}$, test set $D_{v}^{aux}$ of the tested origin $v$, feature extractor $\Featurize$, equivalent relationship $\sim$ for considered origin type, bag size $b$ 
    }
    \Output{membership prediction $m$ of origin $v$ in $D^{train}$}
    
    
    $\mathcal{D}^{proxy} \leftarrow D^{proxy}/\sim$
    
    
    $\mathcal{D}^{proxy, (m)}, \mathcal{D}^{proxy, (n)} \leftarrow \InterSplit (\mathcal{D}^{proxy})$ \tinycomment{Mem/non-member origin for $f^{shadow}$ }\label{line:mem-shadow}
    
    $\mathcal{D}^{proxy, (m)}_{t}, \mathcal{D}^{proxy, (m)}_{n} \leftarrow \IntraSplit (\mathcal{D}^{proxy, (m)})$ \tinycomment{Training/non-training data for $f^{shadow}$}\label{line:train-shadow}

    $f^{shadow} \leftarrow \TrainShadow(f_{\theta}, \mathcal{D}^{proxy, (m)}_{t})$
    
    $S \leftarrow \varnothing$ \comment{Initialize the training dataset for meta-model}

    \For{$D_t \in \mathcal{D}^{proxy, (m)}_{n} \cup \mathcal{D}^{proxy, (n)}$}{
        \If{$D_t \in \mathcal{D}^{proxy, (m)}_{n}$}{
                $S\leftarrow S \cup \GenData(\Featurize, h_i(f^{shadow}, D_t), b, 1)$ \tinytinycomment{Member of $f^{shadow}$}
                \label{line:meta-pos}
            }
        
        \Else{
            $S\leftarrow S \cup \GenData(\Featurize, h_i(f^{shadow}, D_t), b, 0)$ \tinytinycomment{Non-member of $f^{shadow}$} \label{line:meta-neg}
        }
    }
    
    $g=\TrainMeta(S)$ \comment{Meta-model training}
    
    $m=g(\Featurize(h_i(f, D_{v}^{aux}))$\comment{Meta-model inference for the membership of $v$} \\
    \Return{$m$}
    
    \caption{Shadow Training for Data Origin Inference}\label{alg:pipeline}
    
    \end{algorithm*}

\paragraph{Data Reconstruction and Membership Inference} Data reconstruction, sometimes also named model inversion~\cite{model-inversion}, is to reconstruct training data samples on pixel-level. Under some restrictions like relatively small batch size, it is possible to reconstruct newly fed samples according to the shared gradient in federated learning~\cite{zhu2019deep} or two snapshots during online learning~\cite{salem2020updates}.
Membership inference attempts to determine whether a data item is used to train the concerned ML model~\cite{MIA}.
There are tremendous works exploring the capabilities of membership inference under different conditions~\cite{hayes2017logan,leino2019stolen,ChoquetteChoo2020LabelOnlyMI,choquette2021label}. 

\paragraph{Set-level Information Inference} A set-level extension of the original version of membership inference has been developed to check user membership in text\cite{Song2019AuditingDP} and speech \cite{Miao2021TheAA}. ``User" here is a special case of data origin defined in our work.
Another kind of set-level information inference is property inference~\cite{melis2019exploiting,orekondy2018gradient,ganju2018property}. It is to infer binary properties about the training set, which may be unrelated to the model's original primary learning task. 
To enable such inference, there needs to be additional data within and without the concerned property to train shadow models. However, our work doesn't need the additional data of the exact data origin when training shadow models.

\section{Preliminaries}\label{sec:pre}

\subsection{Data Origin}\label{subsec:origin}
Let $e$ be  {\it data generator}, the entity that the data is physically from, e.g. the mobile users who take pictures of their interest, the authors that write text reviews to an online review website. 
Let $c$ be {\it data subject}, the key concept used for data collection of specific usage, e.g. ``movie" in a movie text review website, ``restaurant" in a restaurant pictures collection, ``county" that organizes census. 

Let $\langle e,c,\sigma \rangle$ be a triple, note $x = r(\langle e,c,\sigma\rangle)$ to represent a data input sample $x$ is generated from $e$ to describe $c$ under the data collection noise $\sigma$.  
Here we assume there is no exact the same data input sample $x$ if any of $e$, $c$, $\sigma$ changes, thus $r$ is an bijective function and can be inverse.

\begin{definition}\label{def:ori}
{\bf Data Origin} Given a data input sample $x$, data origin $v$ is either data generator $e$ or data subject $c$, where $\langle e,c,\sigma \rangle = r^{-1}(x)$.
\end{definition}

We don't differentiate $e$ and $c$ when talking about data origin because, for ML developers, it is usually unclear whether the data generator or data subject leads to ML model failure.  

\subsection{Data Origin Inference}

According to Definition~\ref{def:ori}, data origin inference is to get $e$ or $c$ according to $\langle e,c,\sigma \rangle = r^{-1}(x)$. In order to implement this process, there needs some prerequisites, and here we use equivalent relationship to bridge Definition~\ref{def:ori} and actual inference process.  

Let $z = (x,y)$ a data, $x \in \mathcal{X}$ and $y \in \mathcal{Y}$, where $\mathcal{X}$ is the input sample space and $\mathcal{Y}$ is the label space.
For the input sample set $X$ in a certain dataset $D=\{(x,y)\}$, we define equivalent relationship $\sim$ as ``have the same data origin'', i.e. given two data input samples $x_1 \in X$ and $x_2\in X$, $x_1\sim x_2$ means $x_1$ and $x_2$ have the same data origin.

In this work, we assume no exact training data can be obtained for origin inference, but it is feasible to obtain an auxiliary dataset from an exact origin. 
Denote the target training input sample set as $X^{\text{target}}$, and the auxiliary set of $v$ as $X^{\text{aux}}_{v}$, we constrain $X^{\text{aux}}_{v}\cap X^{\text{target}}=\varnothing$. 
In addition, we assume {\it the ML developer has white-box access to the target model $f$, including the model structure, parameters and hidden layer output of any given data input samples}. Denote $h_i(f, X')$ as the hidden layer output of the $i$th hidden layer of $f$, with in input sample set $X'$, and we name this hidden layer as {\it referenced layer}.

\begin{definition}{\bf Data Origin Inference}\label{def:infer}
is to infer the membership of an origin $v$ in the target training set $D^{\text{target}}$, with an auxiliary set $X^{\text{aux}}_{v}$. 
Let $X^{\text{comb}}=X^{\text{aux}}_{v}\cup X^{\text{target}}$.
Taking the $i$th hidden layer output as evidence, $v$ is inferred as involved (member) data origin in $D^{\text{target}}$ only if 
{
\begin{align}\label{eq:infer}
\mathbb{P}\left[ \exists X_k \in X^{\text{comb}}/\sim, X^{\text{aux}}_{v} \subset X_k \Big|  h_i(f,X^{\text{aux}}_{v}) \right ] \geq \delta,
\end{align}
}where $\mathbb{P}$ is the probability and $\delta$ is the chosen threshold depending on the real-world application requirements.

\end{definition}
$X^{\text{comb}}/\sim$ is the quotient set of $X^{\text{comb}}$ regarding $\sim$, which indicates partitioning input data samples according to the concerned origin type. Equation (\ref{eq:infer}) means there exist input samples in the target training set that belong to the same origin as $X^{\text{aux}}$. 

\section{Methodology}\label{sec:label}

\subsection{Overview}
The key technical problem for data origin inference is that {\bf how to determine data origin membership in the training set}. 
This problem is similar to Membership Inference (MI), thus we mitigate their core technique - {\it shadow training} to our solution. 
However, we assume that there are no exact same input samples in the training set when inference, which is different from that in MI. 
Therefore we adopt a vallina variant of {\it embedded-space multiple instance learning} to bridge this gap.

\begin{table*}[t]\footnotesize
\centering
\caption{{\bf Summary of use cases and highest inference accuracy.} In the dataset of the three modalities (image, text and tabular), there are overall six kinds of data origin: mobile user, text author, business, restaurant, movie and county, which are chosen regarding the datasets' purpose and available description (i.e. metadata). ``Mobile user'' and ``Text author'' have the highest accuracy in all kinds of data origins. The text data has the most significant accuracy difference regarding the last layer and the rest of the layers, while tubular data has the smallest.}
\label{tab:sum-exp}
\begin{tabular}{@{}lllrllc@{}}
\toprule
\textbf{Modality} & \textbf{Target Task} & \textbf{Origin Type} & \textbf{Origin\#} & \textbf{Dataset} & \textbf{Target DNN} & \textbf{Accuracy} {\it {\scriptsize Highest / Last layer (Gap)}} \\ \midrule
\multicolumn{1}{l}{\multirow{2}{*}{Image}} & \multicolumn{1}{l}{\multirow{2}{*}{\begin{tabular}[l]{@{}c@{}}Multi-label Classification\end{tabular}}} & \multicolumn{1}{l}{Mobile user} & \multicolumn{1}{r}{823} & \multicolumn{1}{l}{OpenImage\tablefootnote{\url{https://opensource.google/projects/open-images-dataset}}} & \multirow{2}{*}{MobileNet V2} & \multicolumn{1}{r}{0.9456 / 0.9183 (0.0273)} \\ \cmidrule(lr){3-5}
\multicolumn{1}{l}{} & \multicolumn{1}{l}{} & \multicolumn{1}{l}{Restaurant} & \multicolumn{1}{r}{582} & \multicolumn{1}{l}{Yelp Restaurant\tablefootnote{\url{https://www.kaggle.com/c/yelp-restaurant-photo-classification}}} & & \multicolumn{1}{r}{0.8219 / 0.7945 (0.0274)} \\ \midrule
\multicolumn{1}{l}{\multirow{3}{*}{Text}} & \multicolumn{1}{l}{\multirow{3}{*}{\begin{tabular}[l]{@{}c@{}}Polarity Classification\end{tabular}}} & \multicolumn{1}{l}{Text author} & \multicolumn{1}{r}{1328} & \multicolumn{1}{l}{\multirow{2}{*}{Yelp Business\tablefootnote{\url{https://www.yelp.com/dataset}}}} & \multirow{3}{*}{Small Bert} & \multicolumn{1}{r}{{\bf 0.9896} / 0.9324 (0.0572)} \\ \cmidrule(lr){3-4}
\multicolumn{1}{l}{} & \multicolumn{1}{l}{} & \multicolumn{1}{l}{Business} & \multicolumn{1}{r}{160} & \multicolumn{1}{l}{} & & \multicolumn{1}{r}{0.9189 / 0.8381 (0.0808)} \\ \cmidrule(lr){3-5}
\multicolumn{1}{l}{} & \multicolumn{1}{l}{} & \multicolumn{1}{l}{Movie} & \multicolumn{1}{r}{1695} & \multicolumn{1}{l}{IMDB\tablefootnote{\url{https://www.kaggle.com/datasets/raynardj/imdb-vision-and-nlp}}} & & \multicolumn{1}{r}{0.8418 / 0.7819 (0.3170)} \\ \midrule
\multicolumn{1}{l}{Tabular} & \multicolumn{1}{l}{\begin{tabular}[l]{@{}c@{}}Income Prediction\end{tabular}} & \multicolumn{1}{l}{County} & \multicolumn{1}{r}{1780} & \multicolumn{1}{l}{OpenCensus\tablefootnote{\url{https://docs.safegraph.com/docs/open-census-data}}} & MLP & \multicolumn{1}{r}{0.8609 / 0.8531 ({\bf 0.0078})}\\ \bottomrule
\end{tabular}
\end{table*}

\subsection{Primary Pipeline - Shadow Training}

The core idea of the original shadow training is to train one or multiple models - shadow models, that might have the same hyperparameters and function as the target model. 
The shadow model can mimic the ``behaviour'' of the target model, i.e. the pattern of model access output when feeding certain data samples. 
With these collected model access outputs, train a meta-model (usually having different hyperparameters as the target model) to learn the mapping from model access outputs to data membership~\cite{MIA}.
The underline insight behind this technique is that ML models tend to have unique behaviours when the input data is from the training set, e.g. have a higher confidence score~\cite{Salem2019MLLeaksMA}.
This difference can be captured by the meta-model.

Following this idea, we build up  the primary shadow training pipeline in Algorithm~\ref{alg:pipeline}, where we only consider one origin $v$ and one shadow model $f^{shadow}$. 
Proxy dataset $D^{proxy}$ is an external dataset that is disjointed with the target set in terms of data and origins, but can support the training process of the shadow model and meta-model. In Algorithm~\ref{alg:pipeline}, there are three sets divided from $D^{proxy}$:

\begin{enumerate}
    \item The exact training set for $f^{shadow}$, which is denoted as $\mathcal{D}^{proxy, (m)}_{t}$ in line \ref{line:train-shadow};
    \item The dataset that is not the exact training set of $f^{shadow}$, but has the same origins as the training set of $f^{shadow}$ (also the positive training data of meta-model $g$), which is denoted as $\mathcal{D}^{proxy, (m)}_{n}$ in line \ref{line:train-shadow};
    \item A dataset with no data sample and origin overlap with the training set of $f^{shadow}$ at all (also the negative training data of meta-model $g$), which is denoted as $\mathcal{D}^{proxy, (n)}$ in line \ref{line:mem-shadow}.
\end{enumerate}
A more detailed clarification of how to split out the above three sets is in Appendix~\ref{app:partition}. 
Herein we have the positive and negative training data for meta-model training in line~\ref{line:meta-pos} and~\ref{line:meta-neg}. However, there is another technical problem - how to convert origin data to the features that our meta-model can process and learn from.
We adopt the embedded-based feature extraction to solve this problem, which is further described in Section~\ref{sec:MIC}.

\subsection{Efficient Feature Extraction}\label{sec:MIC}
Inspired by embedded-space multiple instance classification, we extract features of the model access data of each origin to enable meta-model training. 
For each origin $v$, we split the model access data into smaller sets - which are called ``bags'' in the context of Multiple Instance Classification (MIC). 
Algorithm~\ref{alg:feat} describes how to split model access outputs into bags and then construct the training data of the meta-model. We adopt a learning-free feature extraction upon each bag to generate the embeddings. The considered feature extraction contains
\begin{enumerate}
    \item {\it Mean and median}: mean and median values along the input data sample matrix axis;
    \item {\it Statistics}: maximum, minimum, mean, $20^{th}$, $25^{th}$, $40^{th}$, $50^{th}$, $60^{th}$, $75^{th}$, $80^{th}$ percentile, variance and standard deviation along axis;
    \item {\it Statistics of Text} (only for text task with last layer as the referenced layer)~\cite{Miao2021TheAA}: mean, maximum and minimum length of the input text, and the difference between the prediction and the ground truth label;
    \item {\it Histogram}~\cite{Song2019AuditingDP}: frequency of the fixed bins of the data samples.
    
\end{enumerate}
The above four statistical feature extraction  are the detailed implementation of $\FuncSty{Feat}$ in line~\ref{line:feat} of Algorithm~\ref{alg:pipeline} and line~\ref{line:gen-feat} in Algorithm~\ref{alg:feat}.
Unlike other learning-based embedding extraction, our approach adapts to high computational costs led by high-dimensional model access outputs.
Furthermore, unlike instance-based MIC, which gives each sample a label, our approach utilizes the collective statistical traits for each bag, avoiding noises induced by a single sample.

\begin{figure*}[t]
	\centering \vskip -0.03in
	\subfigure[MobileNet V2 on OpenImage]{\label{fig:oi-black}\includegraphics[trim={0.0in 0.1in 0.0in 0.1in},clip,width=0.29\textwidth]{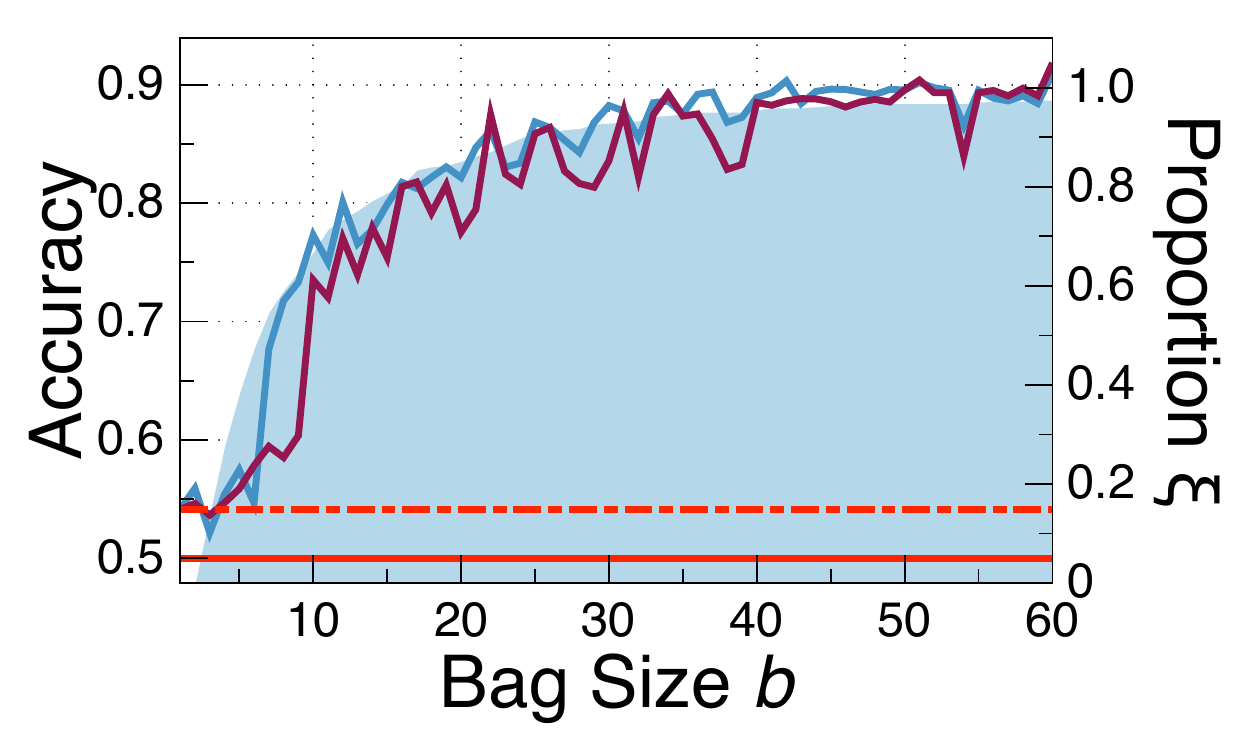}} 
	\subfigure[Small Bert on Yelp Business]{\label{fig:yelp-black}\includegraphics[trim={0.0in 0.1in 0.0in 0.1in},clip,width=0.29\textwidth]{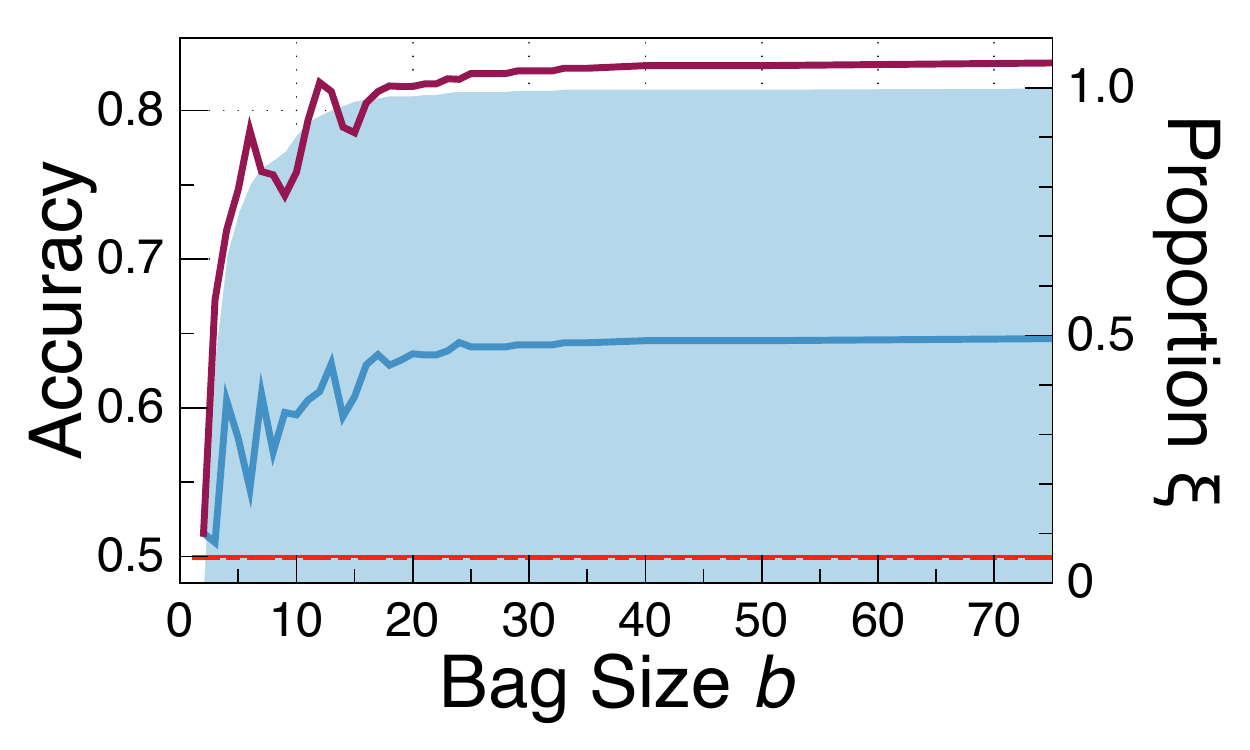}}
	\subfigure[MLP on OpenCensus]{\label{fig:cbg-black}\includegraphics[trim={0.0in 0.1in 0.0in 0.1in},clip,width=0.41\textwidth]{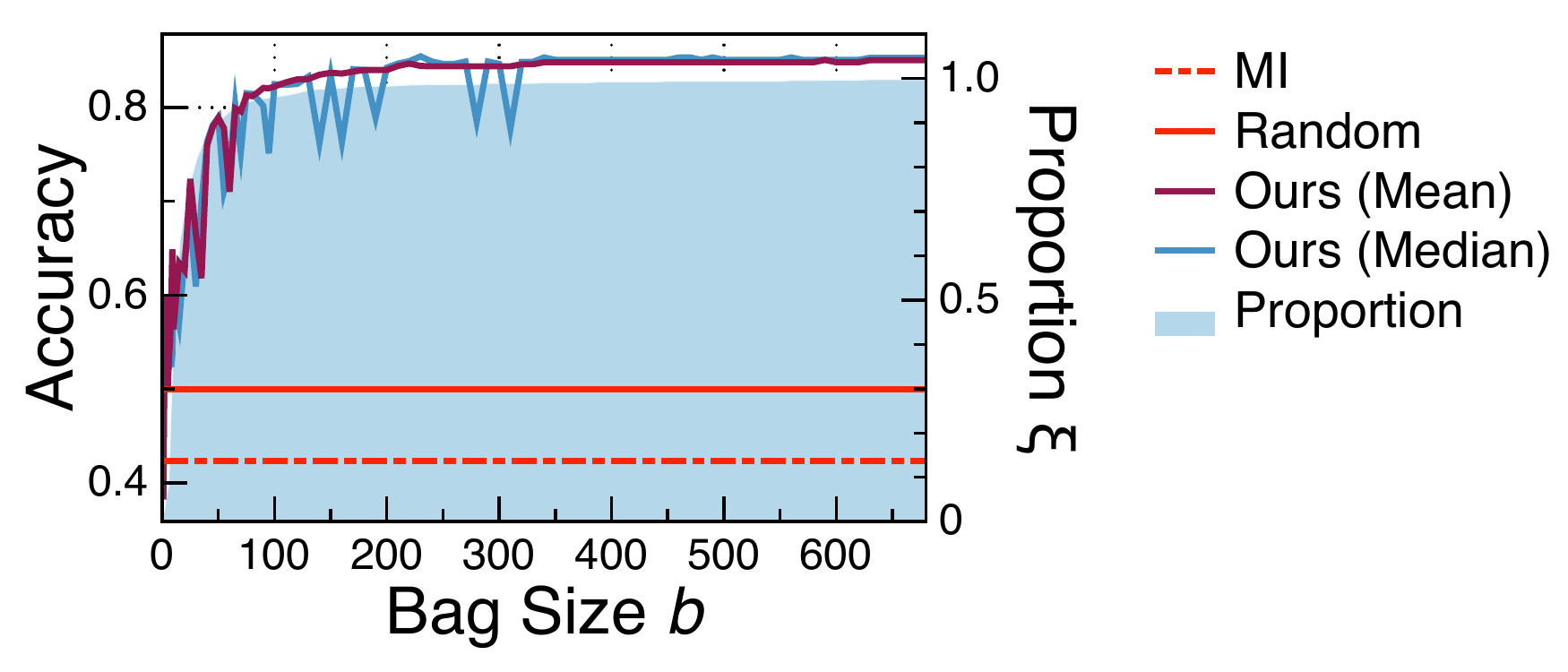}} \vskip -0.1in
	\caption{{\bf Inference accuracy of image, text and tabular data with the last layer as referenced.} $\xi$ is the coverage of origin data, defined by Equation (\ref{eq:cum}) in Appendix~\ref{app:cum}. When $\xi$ increases, there is richer information for each origin and the inference accuracy increases accordingly. However, the increasing accuracy of the three modalities has sensitivity differences.}\vskip -0.18in
\label{fig:black}
\end{figure*}

\begin{algorithm}[t]
\SetKwInOut{Input}{Input}
\SetKwInOut{Output}{Output}
\SetKwFunction{Split}{Split}
\SetKwFunction{GenData}{GenData}

\Input{feature extractor \Featurize, model access outputs $Z_v$, origin membership label $m$, bag size $b$ 
}
\Output{set $S$, which contains the embeddings and the corresponding membership label }

$S\leftarrow \varnothing$

\If{$|Z_v|\le b$}{

$S\leftarrow \{(\Featurize(Z_v),m)\}$ \label{line:feat}
}
\Else{
    $n= \lceil \frac{|Z_v|}{b}\rceil$
    
    $\mathcal{Z}^{'}=\Split(Z_v, n)$ \comment{Partition the elements in $Z_v$ evenly into $n$ sets}
    
    \For{$Z_t \in \mathcal{Z}^{'}$}{
     $S\leftarrow S\cup {(\Featurize(Z_t), m)}$ \label{line:gen-feat}
    }
}
\Return{$S$}

\caption{Generate Meta-model Data $\text{GenData}$}\label{alg:feat}

\end{algorithm}

\section{Experimental Setup}\label{sec:exp}
There are four focal points in our experimental evaluation: 1) what are our use cases (i.e. target tasks and origins); 
2) how to measure our inference performance; 
3) the baseline and devices.

\subsection{Use Cases}

Our use cases are summarised in Table~\ref{tab:sum-exp}. Despite the following, a more detailed experimental setup for each use case is in Appendix~\ref{app:usecase}. 
\paragraph{Users \& Restaurants in Mobile Image Classification}
The training set is an image dataset consisting of pictures taken by mobile users, and the target model is an image multi-classifier. There are two kinds of origins here - ``mobile user'' (data generator) and ``restaurant'' (data subject).
\paragraph{{Authors \& Businesses in Review Polarity Classification}}
The training set is a text dataset consisting of business reviews from website users.
The target task is sentiment polarity classification, which is to determine if a review is positive or negative. 
There are two kinds of origin here: ``text author'' (data generator), and ``business'' (data subject).
\paragraph{Movies in Review Polarity Classification}
The target set is a text dataset containing IMDB reviews towards movies. The target task is still polarity classification and 
``movie'' (data subject) is the origin.
\paragraph{Counties in Income Prediction}
The target set is a structured census dataset.
The target model is an average income predictor, with 1,108 demographic and social characteristics (e.g., age, gender, race, employment) as inputs and implemented by a 20-layer MLP.
The concerned origin is ``county''.

\subsection{Evaluation Metric}
The initial purpose of this work is to identify the training data gap in the deployed ML model on the data origin level. Here ``gap'' and ``non-gap'' share the same importance in real-world applications, as clarified in Section~\ref{sec:intro}. Thus we use {\it accuracy} to measure the inference performance. 

\begin{figure*}[t]
	\centering 
	\subfigure[``Mobile user'' inference on image (OpenImage) dataset.]{\label{fig:oi_white}\includegraphics[trim={0.0in 0.1in 0.0in 0.1in},clip,width=0.48\textwidth]{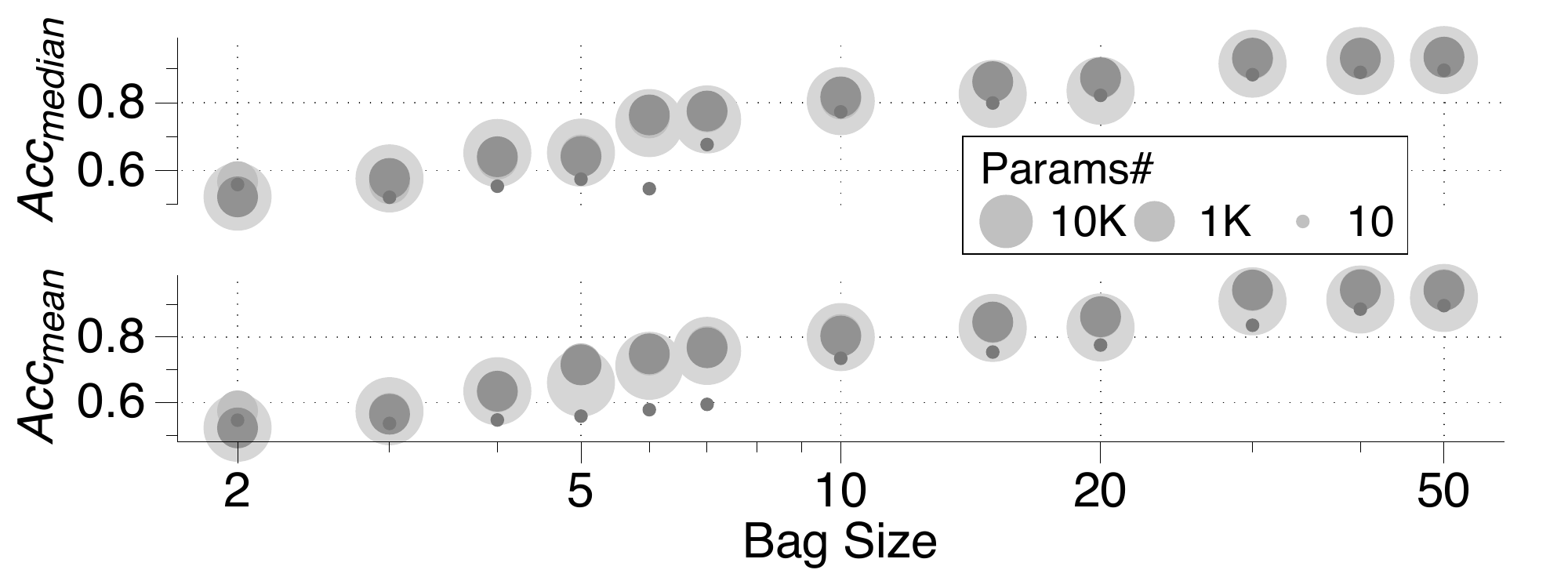}} 
	\subfigure[``Text author'' inference on text (Yelp Business) dataset.]{\label{fig:yelp_white}\includegraphics[trim={0.0in 0.1in 0.0in 0.1in},clip,width=0.48\textwidth]{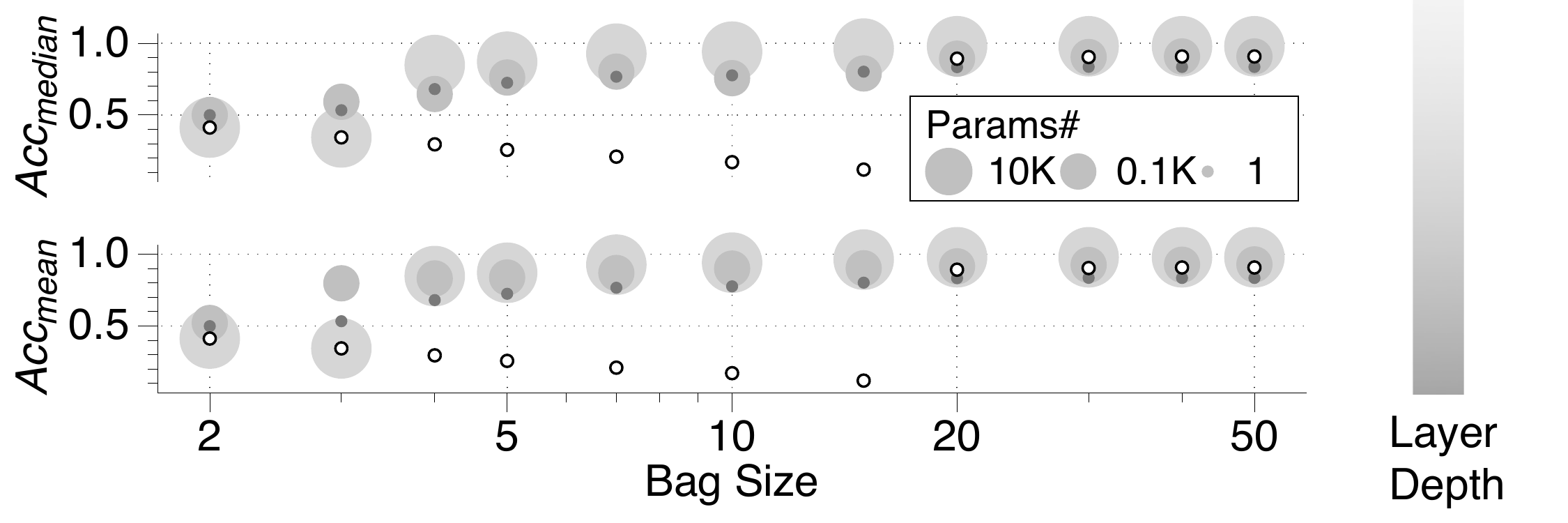}}\vskip -0.1in
	\subfigure[{\it Mean} for ``county'' inference in MLP.]{\label{fig:cbg_white_mean}\includegraphics[trim={0.0in 0.1in 0.0in 0.1in},clip,width=0.48\textwidth]{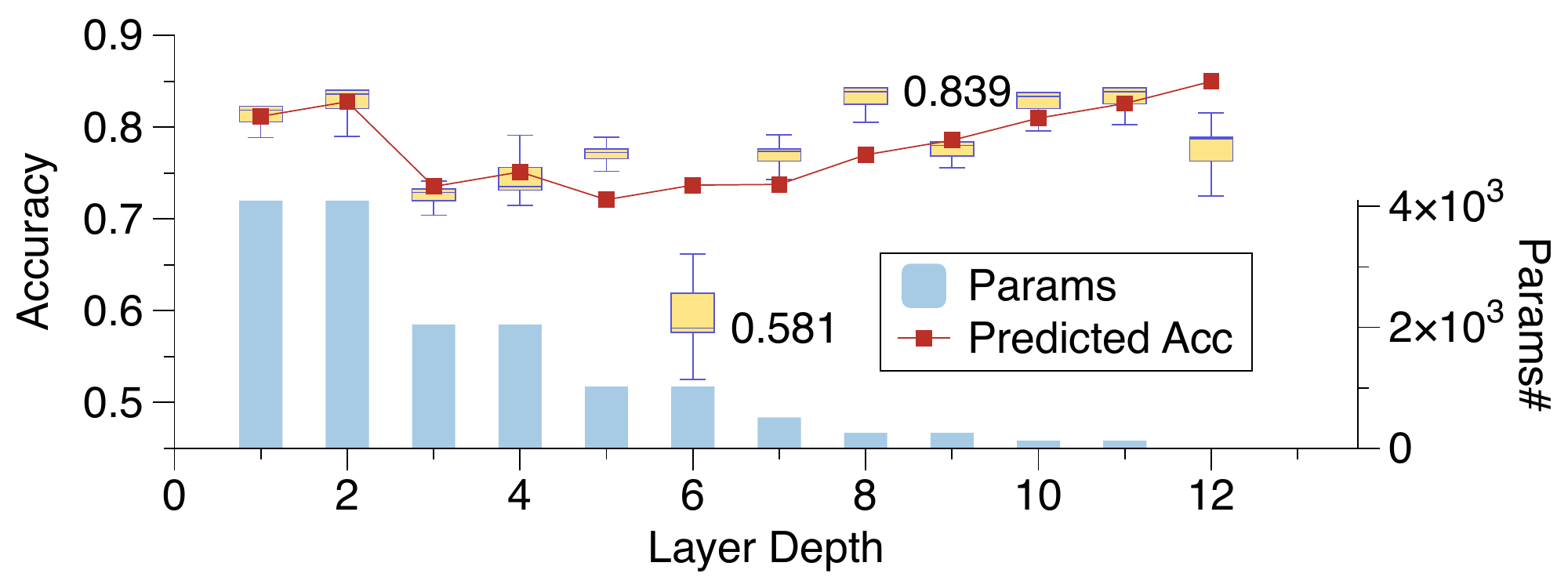}} 
	\subfigure[{\it Median} for ``county'' inference in MLP.]{\label{fig:cbg_white_median}\includegraphics[trim={0.0in 0.1in 0.0in 0.1in},clip,width=0.48\textwidth]{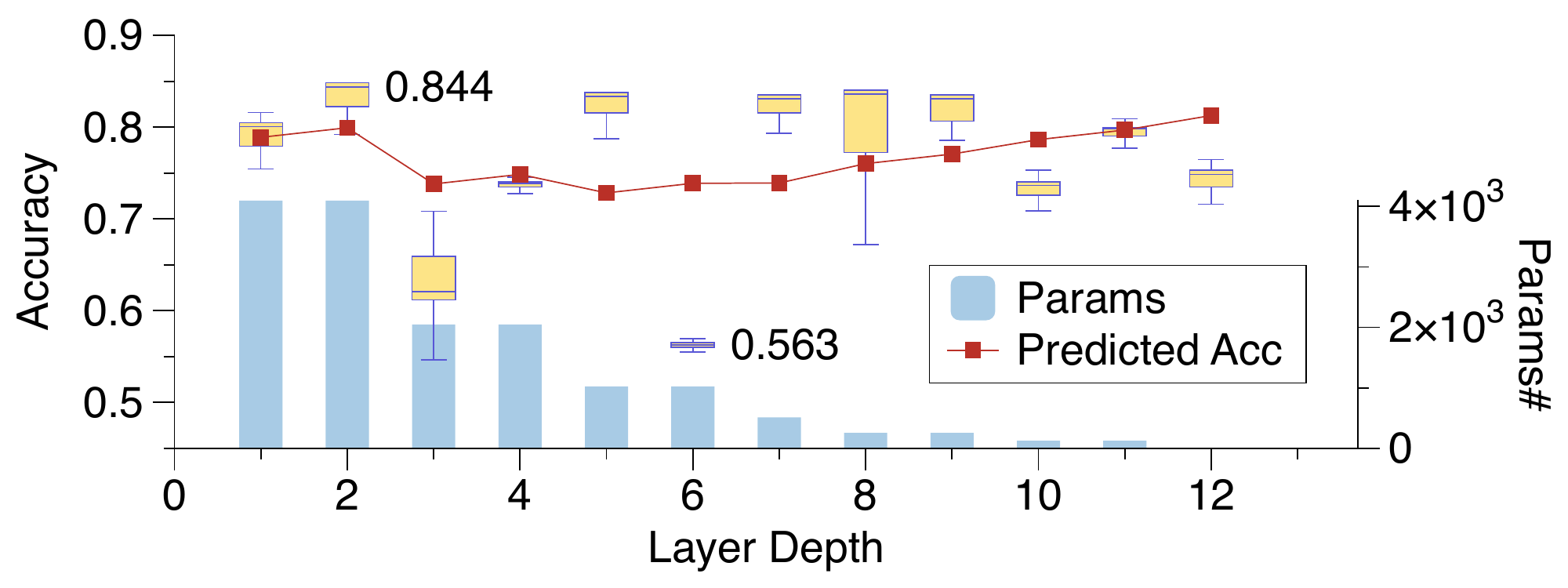}}\vskip -0.1in
	\subfigure[Inference with different layer types in the language model.]{\label{fig:featurization_white}\includegraphics[trim={0.0in 0.1in 0.0in 0.1in},clip,width=0.48\textwidth]{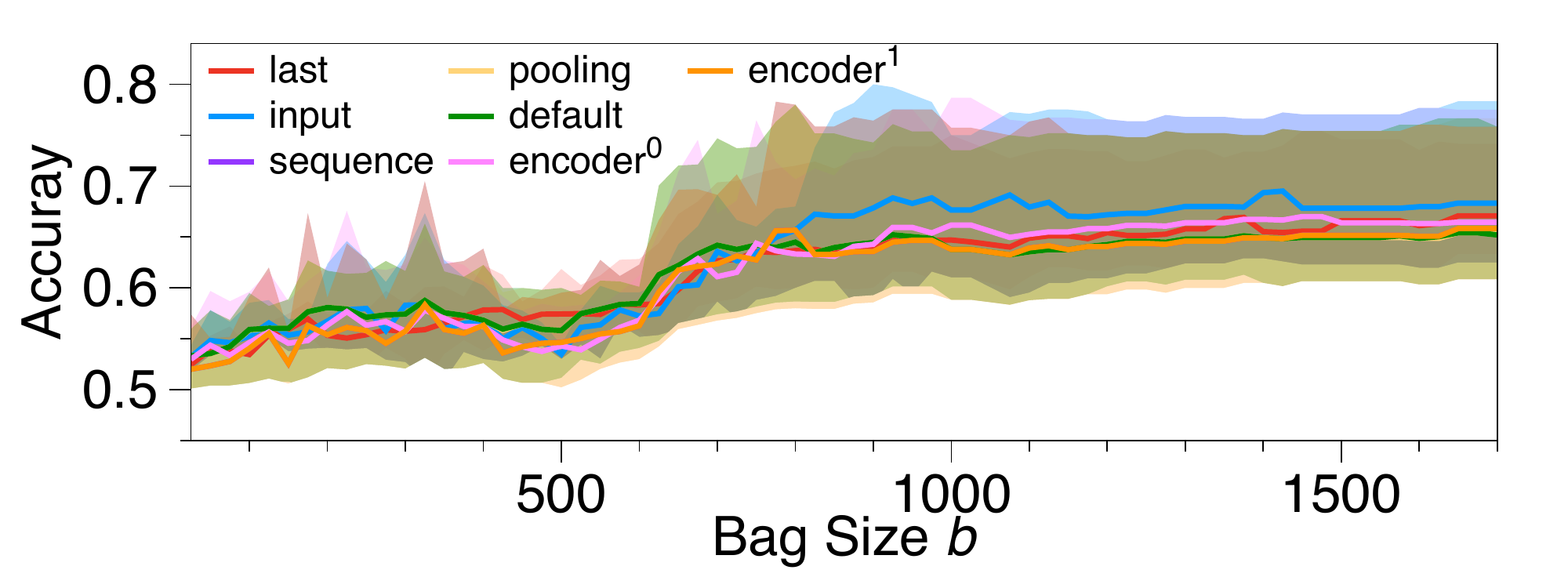}} 
	\subfigure[Training $f^{shadow}$ incrementally from $f$.]{\label{fig:yelp_white_retrain}\includegraphics[trim={0.0in 0.1in 0.0in 0.1in},clip,width=0.48\textwidth]{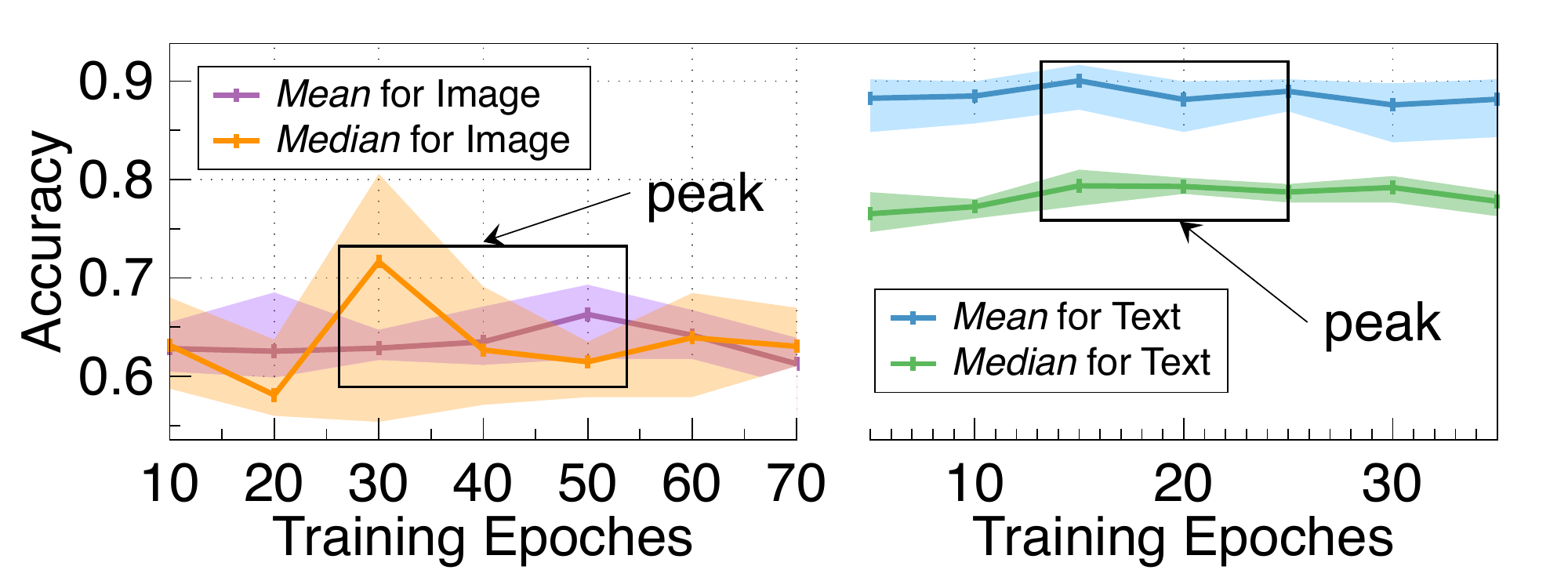}}

	\caption{{\bf Impacts of layer depth, parameters, and featurization.} In (a) and (b), the core functional layers (i.e. convolutional layers in the vision model and encoding layer in the language model) are selected for comparison. The colour shade of the circles indicates the layer depth. In (c) and (d) {\it mean} and {\it median} features are used for inference, respectively. Mostly the layer parameters, layer depth and bag size are positively correlated with inference accuracy, as shown in (a)-(d). However, the inference performance of the language model indicates that the input layer contains the most data origin membership information across the whole network, which can also be seen in (e). In (f), if the shadow model is trained incrementally from the target model, and the concerned data origin is humans (e.g. ``mobile user'' in the image task and ``text author'' in the text review ask), the accuracy first rises and then declines as the retraining epochs increase.}
\label{fig:res}
	
\end{figure*}

\subsection{Baselines and Devices.}

We adopt two baselines for comparison: 
1) Random Guess (abbreviated as \textit{Random}): randomly determine whether the data of the concerned origin is used in the training set;
2) sample-level membership inference (abbreviated as \textit{MI}): we set the bag size as $1$ and implement the standard MI approach, i.e., if a data point is predicted as the member in the training dataset, its origin is used for training.
All experiments are implemented on a Ubuntu 16.04 server equipped with 4 12GB TITAN X Pascal GPUs, 12 Intel Core i7-5930K @ 3.50GHz CPUs, and 62.8GB memory. Besides, all our implementations are based on \cite{tf} 2.5.

\section{Results}

\subsection{Accuracy Overview}
The overall highest accuracy of each use case is shown in Table~\ref{tab:sum-exp}.
For all the use cases, the accuracy of the last layer is lower than that of the intermediate hidden layers. Though we assume the ML developer has white-box access to the target model, it is worth noting that the last layer output has fewer data dimensions, thus when using the last layer as the reference to inference, there is less computation cost.
With this in mind, we get the accuracy gap between the last layer and the intermediate hidden layers with the highest accuracy.
Among the three modalities, the text data has a relatively more significant accuracy gap (over 5\%), while the gap for tabular data is the smallest, which is less than 1\%. 
This gap difference indicates that it is not reasonable for text data to use the last layer as the referenced layer, since the language model suffers a significant accuracy decline when using the last layer output for origin inference.

When using the last layer as the inference reference, we have an analysis of bag size, as shown in Figure~\ref{fig:black}.
The accuracy increases as the bag size increases.
To better understand what bag size means for each origin, we define data coverage $\xi$ in Appendix~\ref{app:cum} to represent the available data integrity of the current bags for an origin.
The area under $\xi$ roughly coincides with the accuracy curve, and this implies that the inference performance is consistent with the integrity of the available (i.e. ML developer can obtain when inference) origin data. 
However, the curve sometimes oscillates, possibly caused by sample noises. This oscillation is larger when the bag size is relatively small ($\leq$10).

\subsection{Impacts of Feature Extraction}
Accuracy comparison of various feature extraction methods in Section~\ref{sec:MIC} is shown in Table~\ref{tab:featurization-black}.
``Compound'' means concatenating all the above features for inference.  
The results show that Statistics and Histogram have higher inference accuracy.

Furthermore, we analyze the performance under various modalities for two basic feature extraction - mean and median. 
For the image data in Figure~\ref{fig:oi_white}, the median is better than that of the mean, yet it is the opposite for the tabular data in Figure~\ref{fig:yelp_white}. 
For the text data in Figure~\ref{fig:black}, the performance of the mean is much better than that of the median. Besides, the median has better performance stability for the image data than the tabular.
The underline reasons might be related to the statistical difference between the mean and median - the mean is more susceptible to global noise than the median, while the median is more locally sensitive.
Among the three modalities, the text task has the smallest input space (the embedding dimension is 128$\times$256), thus its embedding is naturally sensitive to the locality. Thus its median performance is significantly poorer than the others.
The input space of the tabular data is the sparsest among the three, sensitive to the locality but relatively robust to global noises. In contrast, the image data is the opposite, leading to a performance stability difference in the median.

\begin{table}[t]\small
\caption{{\bf Impacts of feature extraction for the text task.} The concerned origin is business in Yelp Business Dataset. The first four kinds of feature extraction are described in Section~\ref{sec:MIC}. ``Compound'' is the combination of all the others. }
\label{tab:featurization-black}
\centering
\begin{tabular}{lr}
\toprule
\textbf{Featurization} & \textbf{Highest Accuracy} \\ \toprule
Mean and median & 0.7760 \\ \hline
Statistics & {\bf 0.8381} \\ \hline
Statistics of Text~\cite{Miao2021TheAA} & 0.7550 \\ \hline
Histogram~\cite{Song2019AuditingDP} & 0.8340 \\ \hline
Compound & {\bf 0.8381} \\ \toprule
\end{tabular}
\end{table}

\subsection{Impacts of Referenced Layers}
There are four factors impacting the inference performance regarding referenced ML model layers - {\it layer depth}, the number of {\it layer parameters}, {\it layer type}, and if the shadow model is trained incrementally from the target model, {\it training epoches}. 

\paragraph{Layer Depth and Parameters} 

A brief comparison of image and text datasets in terms of layer depth and parameters, with origin ``mobile users'' (image task) or ``text authors'' (text task), is shown in Figure~\ref{fig:oi_white} and~\ref{fig:yelp_white}. 
The circle size represents the number of parameters, and the colour shade of the circles indicates the layer depth.
For the vision model, the layer depth dominates the inference accuracy.
However, for the language model, the dominant factor changes to the number of parameters. 
For the language model, we use hollow circles to represent preprocessed embedding input. 
Notably, when bag size reaches a certain threshold, the inference taking the preprocessed embedding as the referenced layer is roughly as accurate as other intermediate layers. 
This implies member and non-member origins are distinguishable in the input sample space of the language model.

We further wonder between layer depth or parameters, which has more impact on the inference accuracy.
We only analyze the case of tabular data, because the language model and vision model modules are highly asymmetrical, which are not suitable for isolated analysis of layer depth or parameters, but MLP is. 
Denote layer depth as $l$ and $|W_{l}|$ as the number of parameters of the $l$th layer, we calculate the Pearson correlation of $(l, Acc)$ and $(|W_l|, Acc)$, respectively. 
For the mean, the correlations are 0.479 and 0.890, and 0.787 and 0.931 for the median. 
Thus, statistically, the accuracy and these two variables are positively correlated, and the correlation between $|W_l|$ and $Acc$ is higher than $l$ and $Acc$.
We further use linear regression models to fit the accuracy scatter, the derived well-fitted curve is shown in Figure~\ref{fig:cbg_white_mean} and ~\ref{fig:cbg_white_median}. For both featurization, the order of magnitude of the linear coefficient for $|W_l|$ is $10^{-2}$ while that of $l$ is $10^{-5}$.
This indicates parameters have more significant impacts on the accuracy than layer depth.

\paragraph{Layer Type}
We only analyze the language model layer types because it has more layer types than the other two.
For the text task with business as the origin, the accuracy of the input layer is significantly higher than that of the other types, which is shown in Figure~\ref{fig:featurization_white}. This finding implies an origin information loss across the Small Bert and is consistent with the text author inference shown in Figure~\ref{fig:yelp_white}.

\paragraph{Training Epoches of the Shadow Model} 
For the proxy model trained incrementally from the target model, as shown in Figure~\ref{fig:yelp_white_retrain}, when epochs are within a certain range, the accuracy is higher than that of proxy models trained from scratch (initialised randomly). However, when the iterations further increase, the accuracy decreases. The excessive training epochs may let the model ``forget''~\cite{9349197} the original information obtained from the target model.

\subsection{Reasoning with Different Origin Types}
Though according to Definition~\ref{def:ori}, data origin is either data generator or subject, the instance of data origin in the real-world application can be interpreted differently from person to person. 
The data origin instance, or the data origin type for a specific application has significant impacts on origin inference, which has been shown in Table~\ref{tab:sum-exp}.
To analyze the impacts of origin type in each use case, we calculate the Pearson correlation coefficient between the origin membership labels and the target task's labels.
With the correlation between the two, we can explore what kind of origin definition is easier to ``memorize''.

The Pearson correlation coefficients of the use cases are in Table~\ref{tab:pearson-corr}. 
For the Yelp Bussiness dataset, there are two kinds of data origin - ``business'', with a correlation of 0.060, and ``text author'', with a correlation of 0.246.
The correlation magnitude relationship is consistent with that of the inference accuracy of ``business'' and ``text author''.
This implies that compared with ``business'', ``text author'' has more impact on the polarity of the review text, thus, the membership information is more easily stored in the intermediate layers of the ML model.  For the image task, the correlation (0.568 $\pm$ 0.403) is overall higher than text (0.060 to 0.246) and structured data (0.533), which can interpret why the inference under the image task outperforms the other two.

\begin{table}[h]\footnotesize
    \centering
 
     \caption{{\bf Statistics analysis regarding origin types.} The Pearson correlation coefficient is calculated by the data origin membership in the training set and the labels of the target tasks. All the correlation coefficients are positive, which indicates that the origin inference performance is positively related to the original task of the deployed ML model itself.}
\begin{tabular}{l|l|l|cc}
\toprule
\textbf{Dataset}                                                         & \textbf{Origin}                                                         & \textbf{Task}                                                                                    & \multicolumn{2}{l}{\textbf{Pearson Correlation}} \\ \toprule
\multirow{8}{*}{OpenImage}                                               & \multirow{8}{*}{\begin{tabular}[c]{@{}l@{}}Mobile \\ User\end{tabular}} & \multirow{8}{*}{\begin{tabular}[c]{@{}l@{}}Multi-class \\ Classification\end{tabular}}           & Paddle                 & Person                 \\
                                                                         &                                                                         &                                                                                                  & 0.830                  & 0.548                  \\ \cline{4-5} 
                                                                         &                                                                         &                                                                                                  & Wheel                  & Clothing               \\
                                                                         &                                                                         &                                                                                                  & 0.357                  & 0.164                  \\ \cline{4-5} 
                                                                         &                                                                         &                                                                                                  & Man                    & Tree                   \\
                                                                         &                                                                         &                                                                                                  & 0.197                  & 0.349                  \\ \cline{4-5} 
                                                                         &                                                                         &                                                                                                  & Building               & Canoe                  \\
                                                                         &                                                                         &                                                                                                  & 0.971                  & 0.526                  \\ \hline
\multirow{2}{*}{\begin{tabular}[c]{@{}l@{}}Yelp\\ Business\end{tabular}} & \begin{tabular}[c]{@{}l@{}}Text \\ Author\end{tabular}                  & \multirow{3}{*}{\begin{tabular}[c]{@{}l@{}}Sentiment \\ Polarity \\ Classification\end{tabular}} & \multicolumn{2}{c}{0.246}                       \\ \cline{2-2} \cline{4-5} 
                                                                         & Business                                                                &                                                                                                  & \multicolumn{2}{c}{0.060}                       \\ \cline{1-2} \cline{4-5} 
IMDB                                                                     & Movie                                                                   &                                                                                                  & \multicolumn{2}{c}{0.014}                       \\ \hline
OpenCensus                                                               & County                                                                  & \begin{tabular}[c]{@{}l@{}}Per Capita \\ Income \\ Prediction\end{tabular}                       & \multicolumn{2}{c}{0.533}                       \\ \bottomrule
\end{tabular}
\label{tab:pearson-corr}
\end{table}

\section{Conclusion}
This paper proposes a novel data origin inference combining shadow learning and embedded feature extraction in multiple instance learning. 
We implement effective inference in three modalities (image, text and tabular). 
We comprehensively investigate the factors from evidence layers, feature extraction, origin definition, etc.
We find that the inference performance is positively correlated with evidence layer depth and parameters.
The future works include 1) further exploration of any other kinds of origin; 2) more efficient inference methodologies that need fewer testing data for the targeted origin.


\bibliographystyle{unsrt}

\clearpage

\appendix
\section{Appendix}\label{sec:appendix}

\subsection{Partition of Origin and Data}\label{app:partition}
This section introduces the data pre-processing pipeline, which supports the shadow training in Algorithm~\ref{alg:pipeline}. A more intuitive pipeline illustration is Figure~\ref{fig:pipeline} in Appendix~\ref{sec:pipeline}.
Initially, there are overall three disjoined datasets:
1) $D^{train}$, used for target model training; 2) $D^{proxy}$, used for training meta model $g$; 3) $D^{extra}$, serving as negative input samples in the testing phase of $g$.
Spotting from ``origin'', there are two kinds of partition during data pre-processing: among data origin (inter-partition) and within data origin (intra-partition).

\paragraph{Inter-origin partition}
For a given use case setting, there are three data origin sets $V^{target}$, $V^{proxy}$ and $V^{extra}$, as shown in Figure~\ref{fig:pipeline} in Appendix~\ref{sec:pipeline}. 
Moreover, there is an origin-level partition inside $V^{proxy}$. This partition is to provide the positive (member) $V_{t}^{proxy}$ and negative (non-member) origin $V_{n}^{proxy,(n)}$ for the meta-model $g$. 
The data of $V_{t}^{proxy}$ is further split out as $\mathcal{D}^{proxy}_{t}$, which is used to train $f^{shadow}$, and $\mathcal{D}_{n}^{proxy, (m)}$, which is then the positive training data of the meta-model $g$. The data from non-member origin $V_{n}^{proxy, (n)}$ are the negative training data of $g$.

\paragraph{Intra-origin partition} 
Intra-origin partition is for the positive training and testing data of the meta classifier $g$. As we assume we don't hold exact training data for inference, the positive data we use is from the member origin of $f^{shadow}$ or $f$ but has no overlap with the exact training data of $f^{shadow}$ or $f$. According to this, the training data is from the member origin of $f^{shadow}$, denoted as $\mathcal{D}_{n}^{proxy, (m)}$; the testing data is from the member origin of $f$, denoted as $\mathcal{D}_{n}^{target, (m)}$.

\subsection{Experimental Details for Use Cases}\label{app:usecase}
\subsubsection{Users \& Restaurants in Image Classification}

In this case, the target training set is an image dataset consisting of pictures taken by mobile users, and the target model is a multi-class image classifier. There are two image datasets in this use case.

The first dataset is extracted from OpenImage, an image dataset that supports diverse visual tasks. 
The target model is an 8-class classifier (i.e. paddle, person, man, tree, wheel, clothing, building and canoe), implemented by popular image backbone on mobile - MobileNet V2~\cite{MobileNetV2}. 
The batch size, the learning rate and the training epochs are 64, 1e-5 and 100, with SGD optimizer with 1e-4 decay. 
For this dataset, we select 823 users with images of such rare transport over ten and split 411 target origins, 370 proxy origins and 42 extra origins.

The second is extracted from Yelp Restaurant, an image dataset for image classification. The target model is a 3-class (i.e. good for dinner, takes reservations and has table service) classifier, the backbone and hyperparameters are the same as that of OpenImage. There are 582 selected restaurants as origins with over 200 images, 133 target origins, 234 proxy origins and 125 extra origins.   

\subsubsection{Authors \& Businesses in Review Polarity Classification}\label{sec:text-yelp}
In this case, the target training set is a text dataset consisting of reviews from website users, and these reviews are for the business.
The target task is review polarity classification, which is to determine whether a review is positive or negative. 
We consider two kinds of origin here: the website user (data generator), and the business (data collection subject).

The experimental datasets are extracted from Yelp Business, a text dataset that contains around 6 million user reviews of 188K businesses. 
Each review is linked to an author ID, stars (further labelled for classification), and the reviewed business ID. 
The target model is implemented by Small Bert\cite{turc2019}. 
We set the batch size to 32, the learning rate to 3e-5 with SGD optimizer with 0.0001 decay, and the training epochs to 100. 

We select 822 active users with over 10 reviews, and partition them into 506 target origins, 780 proxy origins and 42 extra origins.

\subsubsection{Movies in Review Polarity Classification}\label{sec:text-imdb}
In this case, the target set is a text dataset containing IMDB users' reviews, which are for movies. These reviews are for the movies. The target task is to review polarity classification, determining whether a review is positive or negative. 
We consider the movie (processing property) as our origin.
The target model has the same structure and architecture as in Section~\ref{sec:text-yelp}. We select 1695 movies with over 100 reviews, and partition them into 374 target origins, 674 proxy origins and 378 extra origins.

\subsubsection{Counties in Income Prediction}
In this case, the target set is a structured dataset of Census Block Group American Community Survey Data (OpenCensus), which has over 8000 fine-grained attributes on the census block group (neighbourhood) level.
The target model is an average income predictor, with 1,108 demographic and social characteristics (e.g., age, gender, race, employment) as inputs and implemented by a 20-layer MLP. 
We set the batch size, the learning rate, and the training epochs to 128, 1e-4 and 100 with SGD optimizer with 0.0001 decay. 
The concerned origin is a county, which is an administrative or political subdivision of a state in the US. We split counties into 356 target origins, 1295 proxy origins and 129 extra origins. 

\subsection{Coverage of the Origin Data}\label{app:cum}
Since each origin has a different amount of data samples, the absolute values of bag size among different origins are not comparable. Thus we define {\it data coverage coefficient} to describe that in a fixed dataset, when the bag size is set as $b$, for each bag of its corresponding data origin, how much available data are used.
{\footnotesize
\begin{equation}
\SetKwFunction{accumulated}{CumPro}
	\xi(b,X^{test}, \sim) = \frac{\sum_{X^{aux} \in |X^{test}/\sim|} \mathbf{I}(|X^{aux}| \le b)}{|X^{test}/\sim|}.
	\label{eq:cum}
\end{equation}
}where $\mathbf{I}$ is the indicator function, $|X^{test}/\sim|$ is the amount of the origins in the test set, and $\sim$ is the equivalent relation to partition origin data, which is defined in Section~\ref{subsec:origin}. When the bag size $b \ge \max(\{X_t|X_t \in X^{test}/\sim\})$, we have $b/|X^{aux}|=1$, which means each bag contains the all available data in the test set.

\subsection{Experimental Pipeline of Shadow Training.}\label{sec:pipeline}
\begin{figure*}[t]
    \centering
    \includegraphics[width=\textwidth]{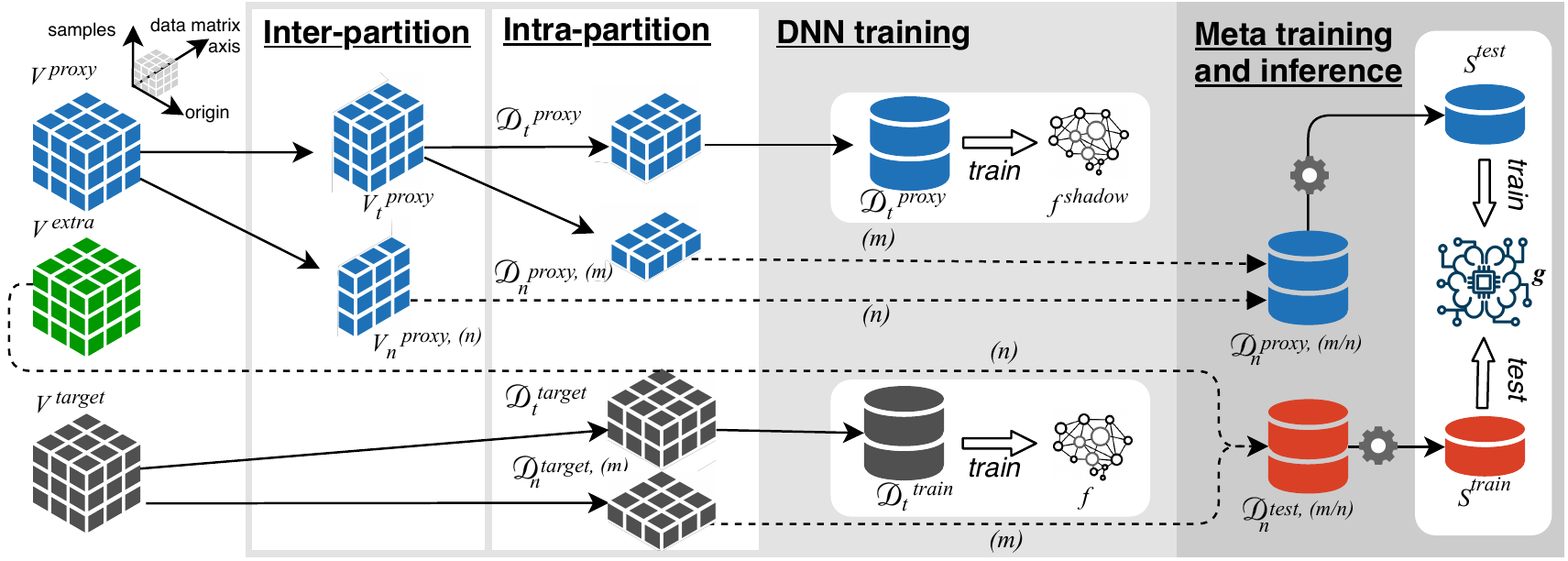}
    \caption{{\bf Experimental pipeline of shadow training.} To support the training of meta-model that finally determine the data origin membership, the available proxy dataset is partitioned according to member/non-member data origin. Then the data of member data origin is partitioned as training/non-training data according to whether it is used to train the shadow model.   }
    \label{fig:pipeline}
\end{figure*}

Figure~\ref{fig:pipeline} shows the data processing pipeline described in Algorithm~\ref{alg:pipeline}.
The three dimensions of the cubes represent data samples, origins and data sample dimensions, respectively.
The data sample dimension is not dividable. Beyond this, in the experiments, we partition origins into different groups ($V_t^{proxy}$, $V_n^{proxy, (n)}$) first, and then for some groups of origins, we further partition data samples ($\mathcal{D}_t^{proxy}$ and $\mathcal{D}_n^{proxy, (m)}$, $\mathcal{D}_t^{target}$ and $\mathcal{D}_n^{target, (m)}$).

\end{document}